\documentclass[10.5pt, conference]{IEEEtran}
\IEEEoverridecommandlockouts
\usepackage{cite}
\usepackage{amsmath,amssymb,amsfonts}
\usepackage{algorithmic}
\usepackage{graphicx}
\usepackage{textcomp}
\usepackage{xcolor}
\usepackage[textsize=tiny]{todonotes}
\def\BibTeX{{\rm B\kern-.05em{\sc i\kern-.025em b}\kern-.08em
    T\kern-.1667em\lower.7ex\hbox{E}\kern-.125emX}}

\begin{document}

\title{Unaligned Sequence Similarity Search Using Deep Learning.\\
}
\author{\IEEEauthorblockN{James K. Senter\IEEEauthorrefmark{1}, Taylor M. Royalty\IEEEauthorrefmark{2}, Andrew D. Steen \IEEEauthorrefmark{3} and Amir Sadovnik\IEEEauthorrefmark{4}}
\IEEEauthorblockA{\IEEEauthorrefmark{1}\IEEEauthorrefmark{4}Department of Electrical Engineering and Computer Science\\
\IEEEauthorrefmark{2}\IEEEauthorrefmark{3}Departments of Microbiology and Earth and Planetary Sciences\\
University of Tennessee - Knoxville\\
Knoxville, TN\\
Email: \IEEEauthorrefmark{1}jsenter3@utk.edu,
\IEEEauthorrefmark{2}troyalty@vols.utk.edu,
\IEEEauthorrefmark{3}asteen1@utk.edu,
\IEEEauthorrefmark{4}asadovnik@utk.edu}}

\maketitle

\begin{abstract}
 Gene annotation has traditionally required direct comparison of DNA sequences between an unknown gene and a database of known ones using string comparison methods. However, these methods do not provide useful information when a gene does not have a close match in the database. In addition, each comparison can be costly when the database is large since it requires alignments and a series of string comparisons. In this work we propose a novel approach: using recurrent neural networks to embed DNA or amino-acid sequences in a low-dimensional space in which distances correlate with functional similarity. This embedding space overcomes both shortcomings of the method of aligning sequences and comparing homology. First, it allows us to obtain information about genes which do not have exact matches by measuring their similarity to other ones in the database. If our database is labeled this can provide labels for a query gene as is done in traditional methods. However, even if the database is unlabeled it allows us to find clusters and infer some characteristics of the gene population. In addition, each comparison is much faster than traditional methods since the distance metric is reduced to the Euclidean distance, and thus efficient approximate nearest neighbor algorithms can be used to find the best match.  We present results showing the advantage of our algorithm. More specifically we show how our embedding can be useful for both classification tasks when our labels are known, and clustering tasks where our sequences belong to classes which have not been seen before.
\end{abstract}


\section{Introduction}
The central dogma of biology states that all organisms contain DNA, which is transcribed into RNA and then translated into proteins, which catalyze the chemical reactions that define life. DNA sequences that encode for specific proteins are known as genes. Thus, understanding the function of DNA sequences that encode genes is a fundamental task across all fields of biology.


In order to interpret sequence data, it is usually necessary to annotate sequences identified as genes. This is commonly done by aligning unknown sequences to ones of known function using algorithms such as Basic Local Alignment Search Tool (BLAST) \cite{altschul1990basic} and comparing them based on the fraction of identical nucleotides (or amino acids, after in silico translation). Sequence identity comparisons are typically very accurate when sequence identity is high, which is one of the reasons these methods are so common in comparative genomics \cite{rost2002enzyme}. However, these methods do have downsides which prevent them from being useful under certain conditions.

First, when using these methods there are still many genes which cannot be annotated.  For instance, in the Tara Oceans data set \cite{pesant2015open}, an average of 50\% and up to 80\% of bacterioplankton genes lacked sufficient homology to genes in databases of known function to be confidently annotated \cite{Sunagawa2015structure}. This should be no surprise: gene function is overwhelmingly studied in genes that derive from microbes that grow in culture, whereas the vast majority of microbes on Earth belong to uncultured taxa  \cite{steen2019high,lloyd2018phylogenetically}.

Second, even when using annotated data-sets, annotations are often vague. For instance, the 100 most common annotations in the RefSeq database of high-quality genomes \cite{oleary2016refseq} include such vague annotations as ‘acyl carrier protein’, ‘porin’, and ‘peptidase’. Any of these broad categories contain an enormous diversity of gene sequences and tertiary protein structures. 

In addition, even high-quality annotations lose important information, because many types of important sequence information, such as relative amino acid content or factors that affect temperature optima of gene products, are discarded during the annotation process. These factors may be an important part of differences in ecosystem function, but they would not show up in ecosystem analyses based on annotations. 

Finally, BLAST searches can be slow, especially when we wish to compare our sequence to multiple others and not just find the best match. This is mainly due to the need for alignment and string comparison.

Here we embed DNA or amino acid sequences in a low-dimensional space with a neural network combining convolutional and recurrent layers. The embedded data can be searched much more rapidly thanks to dimensionality reduction. Furthermore, the convolutional layers of the network allow recognition of important features while remaining robust to sequencing error such as insertions or deletions, and the LSTM layers capture long-term correlations in sequences that may be biologically important but difficult to identify in sequence alignments. Finally, this approach can identify sequences that are similar in some biological respect but which may not have any measurable sequence alignment, an ability which may be useful in sequence analysis tasks other than sequence annotation, such as identifying properties of gene products like temperature optima or enzyme lifetime.

\begin{figure}\centering
\includegraphics[width=\linewidth]{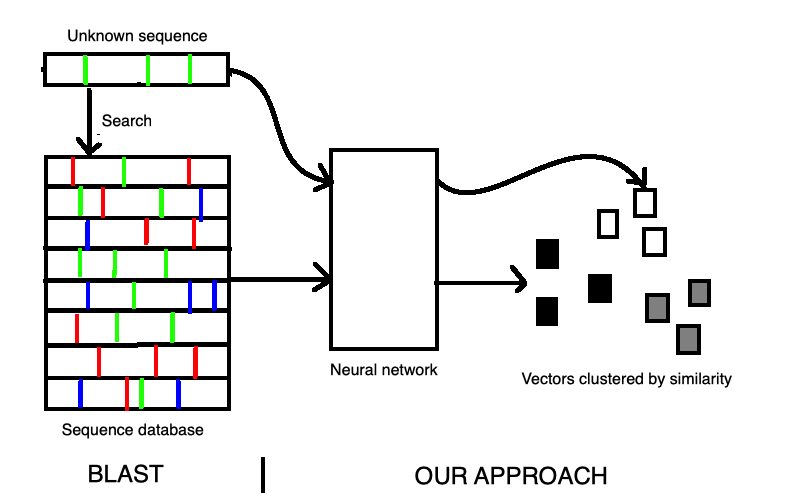}
\caption{In BLAST (left), a database of proteins must be searched for those with similar seeds (short strings) to the unknown protein. In our approach (right), the known proteins and the unknown protein are all transformed by a neural network into a low-dimensional vector representation where proteins can be grouped by similarity.}
\label{fig:blast}
\end{figure}

Our paper is structured as follows. In Sec. \ref{sec:related} we discuss the previous research on which our work is based. This includes both work in bioinformatics regarding DNA sequence interpretation, and recent machine learning research on sequence classification and embedding space learning. In Sec. \ref{sec:method} we motivate and describe the network architecture we use, and the training method which allows us to learn a useful embedding space. In Sec. \ref{sec:results} we present results on multiple different experiments, and analyze some of the parameters in order to choose the ideal ones. Finally, we conclude in Sec. \ref{sec:conclusion}.

\section{Related Work}
\label{sec:related}
\subsection{DNA Sequence Interpretation}

BLAST and its variants (e.g. gapped BLAST and PSI-BLAST \cite{altschul1997gapped}, BLAST+ \cite{camacho2009blast+}) have traditionally been used to identify regions of sequence homology between a query sequence and database of reference sequences. The algorithm includes 3 main steps. (1) A list is compiled of important seeds (short strings of nucleotides or amino acids) appearing in the query sequences. (2) The reference database is scanned to find locations of of the same seeds, aided by an index of seed locations. (3) Matches between query seeds and reference seeds are extended to determine whether the areas neighboring the seeds match as well as the seeds. For each query sequence, the reference sequences with the best matches are returned.

Several improvements have been made to the original BLAST. USEARCH \cite{edgar2010usearch} reduces search time by returning only a few high-quality matches rather than considering all possible matches. DIAMOND \cite{buchfink2015fast} constructs a double index to traverse query and reference seeds more quickly.  GPU-BLAST \cite{vouzis2010gpu}, HPC-BLAST \cite{sawyer2015hpc}, and H-BLAST \cite{ye2017hblast} parallelize the database search on high-performance systems.

BLAST and its improvements all have the same search limitation: They search for very close matches using certain confidence levels, and do not provide distances to the entire set (or subset). Comparing to all sequences would be very expensive since it would require multiple alignments and string comparisons. Our goal is to allow faster comparisons by providing a significantly smaller numeric representation of each protein in the database by preprocessing it with our neural network. Euclidean distance on short vectors is a faster similarity metric than string comparison on long sequences. Also, clustering-based algorithms such as fast nearest neighbors \cite{muja2009fast} or approximate nearest neighbor search algorithms such as Neighborhood Graph and Tree \cite{iwasaki2018optimization}  allow for nearest neighbor search of the database in sub-linear time. Figure \ref{fig:blast} illustrates how our approach differs from BLAST.

As a practical matter, Hidden Markov Models (HMMs) are often part of gene annotation strategy, \cite{eddy2011accelerated}. For instance, the popular annotation package PROKKA \cite{seemann2014prokka} uses a hierarchical strategy, beginning with BLAST+ searches of increasingly expansive databases and ending with HMM searches of protein family databases, e.g. \cite{haft2012tigrfams}. However, these have their own limitations as well. Mainly, since the DNA sequence is assumed to be Markovian, it means that the state transitions depend only on the current state and not anything in the past. This is most likely not physically true for genetic data, and although we can alter the HMM’s to consider previous states as well, we must define exactly how many previous states will be included. 

Our approach is to embed sequences in a lower-dimensional space, and to use those vectors for comparison. This has been done in a few previous works. One technique to embedding a DNA sequence is \cite{woloszynek201916s}, which uses a word2vec model to encode short sequences. DSPACE \cite{schwartz2018deep} is more similar to our approach, training a neural network to embed amino acid sequences for multiple supervised tasks. Unlike these models, we introduce LSTM layers which allows dealing with sequences of different lengths in addition to capturing dependencies which are distant in the sequence (as opposed to convolutional layers which only capture local patterns) . Our network structure is inspired by DanQ \cite{quang2016danq}, which uses 1-dimensional convolutional layers followed by LSTM layers. The convolutional layers recognize short-term patterns, while the LSTM layers recognize long-term patterns, making the combination stronger than either half alone. We adapt this structure by adding bidirectionality and the possibility to deal with varying length sequences to produce embeddings and show that this network is more accurate than DSPACE, which only includes convolutional and dense layers.

\subsection{Deep Learning on Sequences}

Machine learning using sequence data is not a new problem and has been studied extensively for many years, particularly in the natural language processing community. One of the early methods used for this type of data was the HMM \cite{1baum1967inequality} which is a statistical model in which the process is Markovian with observable (the signal) and hidden (the prediction) events. Machine learning can be used to find the HMM parameters and dynamic programming (for example the Viterbi algorithm \cite{2viterbi1967error}) can then be used to find the maximum likelihood predictions. These methods have been used for many different types of sequence data such as speech recognition \cite{3rabiner1989tutorial} and gene finding \cite{4lukashin1998genemark}.

More recently, with the advent of deep learning algorithms, a new set of machine learning algorithms has been developed which is not limited by the same constraints and therefore is able to achieve much better results. More specifically, recurrent neural networks (RNNs) have been able to achieve excellent prediction results on sequences since they can utilize high dimensional hidden states which can remember an unlimited amount of past information \cite{5sutskever2011generating}.  This allows the network to discover much longer temporal dependencies as compared to HMM’s. One type of  RNN, the long short-term memory network (LSTM) \cite{6hochreiter1997long}, has been especially successful in achieving state of the art results on a variety of tasks as it is able to better learn long term dependencies \cite{7graves2013generating}. Finally, bidirectional RNN's \cite{schuster1997bidirectional} have been used for prediction in both the forward and backward directions of the sequence.

In this work we adopt these methods to work on DNA sequences. More specifically, we use bidirectional LSTM's on top of convolutional layers to predict the protein class of an unknown gene.

\subsection{Learning Embedding Spaces}

The process of training a neural network on one task and using an intermediate layer of that network to create an embedding space for different tasks has been applied in different domains. For example, Word2vec \cite{mikolov2013efficient} learns a vector representation of words by training a network to predict future words from past words, and the resulting embedding places words with similar meaning closer together. 

VGG-Face \cite{parkhi2015deep} creates an embedding of faces, for use in face recognition: learning the distinguishing features of each face regardless of position, lighting, etc. As not all identities are known at training time it is not possible to build a simple face classifier, and therefore the task is to determine if two faces are of the same person. This is accomplished by learning an embedding in which faces of the same person are close to each other, while faces of different people are far in the embedding space. They use two different types of training: either using triplet loss to train the embedding directly, or training a classifier and removing the final layer to get the embedding. In our work we adopt the latter.

The concept is the same across domains: a high-dimensional input space is converted to a low-dimensional space containing the most important features of each element for a specific task. The distance between two elements in this low dimensional space is a measure of their similarity, useful for many tasks beyond the original training task. For example Word2vec vectors can be used for sentiment analysis \cite{ZHANG20151857}, while VGG-Face embeddings can be used to search for lookalikes \cite{sadovnik2018finding}.

In this work we use deep neural networks to learn an embedding space for DNA sequences. Similar to VGG-Face \cite{parkhi2015deep} we assume that we don't know all protein classes ahead of time and therefore cannot rely on a classifier. In addition, as has been shown, we expect these embeddings to be useful for other tasks as well. Although we use a similar training setup to VGG-Face, our network architecture is different than theirs as we are dealing with sequence data and therefore use RNN's. 

\section{Methods}
\label{sec:method}

We frame the original problem we are trying to solve in the following way. Given a query DNA sequence $x_q$, we wish to label it with a protein label $y_q$ (i.e. N-acetyltransferase) . In addition, we have a database of $N$ other DNA sequences $X_d=[x_{d1},x_{d2}...x_{dN}]$, each with its own label $Y_d=[y_{d1},y_{d2}...y_{dN}]$. We wish to compare $x_q$ to all sequences in $X_d$ in order to find a match, and transfer the label.  For example, if the best match to our query sequence $x_q$ is $x_{dm}$, we can simply give it the label $y_q=y_{dm}$. This matching process can be done using algorithms such as FASTA \cite{pearson2003finding} and BLAST \cite{altschul1990basic} which provide a matching score between two sequences.

However, as these algorithms require alignments and string comparisons they can only return a few best matches. In addition, these algorithms have no direct way to measure the importance of certain subsequences. Therefore, in this work we propose a different way to compare the sequences. We first learn an embedding method $f(x;\theta)\in \mathbb{R}^d$. The function embeds a varying length sequence into a $d$-dimensional Euclidean space. Once we have learned such an embedding, the similarity between two sequences (regardless of their length) can be found simply by calculating their Euclidean distance $||f(x_q;\theta)-f(x_{dm};\theta)||$.

The function's parameters $\theta$ can be learned using our labeled database. The goal of learning would be to create a space where genes with similar function are close to each other as compared to ones with different function. That is:
\begin{equation}
\label{eq:triplet}
    ||f(x_a;\theta)-f(x_b;\theta)|| < ||f(x_a;\theta)-f(x_c;\theta)||\\
\end{equation}
 Where $x_a$ and $x_b$ are both labeled with $y_a$ and $x_c$ is not labeled with $y_a$. Learning a space in which Eq. \ref{eq:triplet} is true gives us the advantage of being able to do more than simply label the sequence $x_q$ using our known database. For example, given distances to multiple other sequences in $X_d$ we can infer something about the functionality of the query sequence. In addition, given a group of unknown genes, we can use the embedding space to cluster them, thus finding out how many types of genes there are and how they relate to each other.  
 
Similarly to other works in computer vision we can do this by first training a deep neural network classifier on a large number of classes using our labeled database $X_d, Y_d$. We can view the output of the network's penultimate layer as the embedding vector $f(x;\theta)$. The final dense layer can be viewed as a linear classifier over the embedding layer, and therefore we expect genes of similar function to be close in the embedding space.

Therefore, although we use the classification layer for training and to test our classification task, it is removed when comparing sequences to one another. By simply calculating the Euclidean distance of the output vectors from the embedding layer we can measure the similarity between sequences. 

\subsection{Network Architecture}
\label{subsec:architecture}

\begin{figure}\centering
\includegraphics[width=\linewidth]{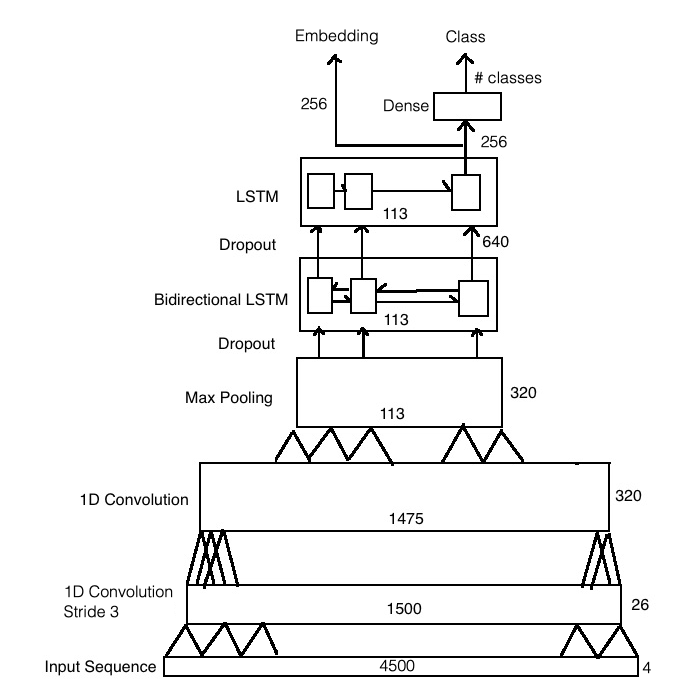}
\caption{Network architecture, including sizes for each layer and output. All layers are presented except for the non-linear activations. Notice that the top layer (Dense) is only used for training and during the classification task. For embedding we remove this layer and directly use the 256 length vector produced by the LSTM. Details are provided in Sec. \ref{subsec:architecture}.} 
\label{fig:network}
\end{figure}

Our network is detailed in Figure \ref{fig:network}. The input to the network is a $4500\times 4$ matrix. The columns represent a one-hot encoding of the 4 nucleotides $(A,C,G,T)$ and the rows represent the maximum length of a sequence. For example, a value of $1$ at position $(0,0)$ means that the first nucleotide in the sequence is $A$ and positions $(0,1), (0,2), (0,3)$ will be zero. If a sequence is shorter than 4500, all the extra columns are set to $0$. Masking is applied so that the zero padding does not affect the final result. 

The first layer is a 1D convolution (stride 3, kernel size 3, 26 filters) to represent encoding of every 3 nucleotides into amino acids. Another 1D convolutional layer (stride 1, kernel size 26, 320 filters) followed by a max pooling layer (stride 13, kernel size 13) reduces the size of the sequence and represents short patterns of amino acids. The sequence then passes through a bidirectional LSTM layer (output size 640), retaining an output for each step of the sequence. This layer captures long-term trends in the sequence regardless of direction. Next, a forward LSTM layer, retaining only the output from the last step, collects a summary of the sequence. The output from the final LSTM layer is the embedding: a 256 length vector. 

For training through classification, the embedding layer is followed by a dense layer with output size equal to the number of classes. Training is performed by minimizing the cross entropy loss using batch gradient decent. Given that the final dense layer is a simple linear classifier, the embedding should learn features that are useful for classification without itself being tied to specific classes. The convolutional layers have ReLU activation, the LSTM layers have tanh activation, and the final class layer has softmax activation. 

For comparison we also trained a DSPACE model, using the same architecture as in the source code of \cite{schwartz2018deep}, with an extra stride 3 convolutional layer after the input to account for using DNA sequences instead of amino acid sequences. The inputs to this sequence are the same: length-4500 sequences with extra space padded with zeros, but masking was not possible as this model requires inputs of a fixed length. This network contains several 1D convolutional layers followed by several dense layers, culminating in an embedding layer. After the embedding layer, we replaced their output layers with a dense layer for class prediction. 

We plan on releasing all of our code and models as part of the publication of this paper. In addition we will release the data-sets we used to train and test the models to ensure reproducibility.

\subsection{Classification Training} 
\label{subsec:training}
We used protein sequences from the RefSeq database \cite{oleary2016refseq}, v83, filtered to contain only bacteria and archaea. Our first training set consisted of approximately 40 million sequences from the 30 most common classes. As an example we show the top 10 class names in Table \ref{tab:classnames}. The training set and the test set had the same proportion of each class. We then collected separate datasets containing the most common 100 and 1000 classes. For better training efficiency, we did not use all sequences from these classes, keeping only about 16 million sequences per dataset with equal representation for each class. Each training dataset had a corresponding test set of approximately 1 million sequences, also with equal numbers of examples from each class.

\begin{table}
\centering
\begin{tabular}{|l|l|}
\hline
 ABC transporter ATP-binding protein & MFS transporter \\
LysR family transcriptional regulator & transcriptional regulator\\
ABC transporter permease & membrane protein\\
DNA-binding response regulator & N-acetyltransferase\\
TetR/AcrR family transcriptional regulator & alpha/beta hydrolase
\\\hline
\end{tabular}
\caption{\label{tab:classnames}The top 10 protein classes in our subset of the RefSeq dataset \cite{oleary2016refseq}. These provide an example of the type of classes we are using for classification.}
\end{table}

We trained our model on each of the 30, 100, and 1000 class datasets, plus a DSPACE model on the 100 class dataset. The models were trained to minimize categorical cross entropy loss using an Adam (for LSTM) or Nadam (for DSPACE) optimizer with learning rate 0.001. Each network was trained on 200,000 random batches of 100 sequences. Training a model on a Quadro P5000 GPU took approximately two days. 

\subsection{Embedding Analysis}
\label{subsec:embedding}
To test the quality of each embedding on unseen classes, we arbitrarily chose 1000/10,000 classes not used for training. From each class we chose a random pair of sequences, and treated the first of each pair as a query $x_q$, while the second sequences of all pairs were the database $X_d$. If the embedding accurately reflects the biological function of the sequence, two sequences belonging to the same class should be closer in the embedding space than two from different classes (Eq. \ref{eq:triplet}). 

We therefore found the Euclidean distance from each query sequence to the entire database, producing a ranking of the most similar genes from the database. Then, we determined how many queries had the correct answer (the sequence from the same class) within the top N closest database sequences, where N = 1, 10, 20, or 50. The fraction of gene pairs placed close together by the embedding (with several definitions of closeness) is similar to the information retrieval measure ''recall-at-n'' and is a way to quantify the embedding, or how well the model can group proteins from classes never seen before.

\section{Results}
\label{sec:results}

\subsection{Classification Results}
\label{subsec:class_results}
We first present the results of using our neural network architecture for classification; that is, both the training set and the test set contain DNA sequences with the same protein labels. The number of classes and division between training and test set are described in Sec. \ref{subsec:training}.

Table \ref{tab:test} shows the accuracy on the test set for a different number of classes. In addition, we compare our results with the DSPACE model \cite{schwartz2018deep}. As expected, as the number of classes increases our results slightly decrease since there is more chance for error. However, even when the number of classes is multiplied by 10 (from 100 to 1000) the accuracy only drops by a few percent showing that the network scales well. Our model clearly outperforms the DSPACE baseline and shows that the network is able to label DNA sequences from classes it has been trained on.

\begin{table}
\centering
\begin{tabular}{l|l}
Model & Test Accuracy \\\hline
30 class LSTM & .968 \\
100 class LSTM & .914 \\
1000 class LSTM & .896 \\
100 class DSPACE & .832
\end{tabular}
\caption{\label{tab:test}Accuracy on test data when training and testing our network with different amounts of classes. We also compare to other previously published network architecture DSPACE \cite{schwartz2018deep}.}
\end{table}

\begin{figure}\centering
\frame{\includegraphics[width=\linewidth]{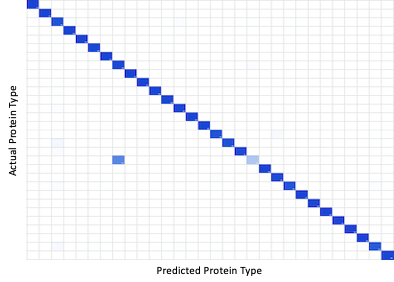}}
\caption{The confusion matrix for our 30 class model. The colors scale from dark blue (100\%) to white (0\%). Notice how all classes are nearly classified perfectly except for one class due to our ground truth labels. More information is given in Sec. \ref{subsec:class_results}  }.
\label{fig:confusion}
\end{figure}

Fig. \ref{fig:confusion} shows the confusion matrix based on our 30 class model. It adds another way to interpret the results, and emphasizes a flaw in the data and labels we are using. Although our total classification rate for the 30 class model is 96.8\%, when looking at the confusion matrix it is clear that most classes achieve a very high precision, with only one class achieving a 32\% accuracy (class 18), which is mostly classified as class 7. However, the labels of these classes are "N-acetyltransferase" (7) and "GNAT family N-acetyltransferase"(18), which are essentially the same protein but are separated into two different classes in the RefSeq dataset. This `error' suggests that the network has correctly learned that two differently named classes refer to the same broad set of sequences.



\subsection{DNA Sequence Embedding Results}
Next we present results to show that our embedding space is meaningful even for DNA sequences which are not from the classes which appeared in the training data. We remove the final classification layer from the network and use the testing strategy discussed in Sec. \ref{subsec:embedding}. 

Tables \ref{tab:embed} and \ref{tab:embed10k} show results of the embedding analysis with pairs of sequences from 1000 and 10,000 classes, respectively. As expected the probability of having the closest sequence be of the same class as the query one is lower than our classification results. However, it is important to look at the entire table to realize the advantages our method provides. For example, when training on 1000 classes the most similar sequence is of the same class 53\% of the time (Table \ref{tab:embed}). When comparing this to a random chance of 0.1\% this is an impressive result especially given that these classes have never been seen by the network and there is only one matching sequence in our database.

In addition, if we do not only focus on the closest sequence, but instead look at the top $N$ sequences, we see that although the correct match is not always ranked the highest, it is usually ranked high. For example, when examining the top 50 classes out of 10,000 (0.5\% of the results) the correct result is there 70\% of the time. This result could be extremely useful since if needed we can then perform a more traditional sequence comparison (for example BLAST) on a much smaller subset of our database and still find the correct match.

A few other observations can be made when analyzing the results. First, as the number of classes used for training increases (30, 100, 1000) so does the accuracy of the embeddings. This is reasonable since the network can generalize better when ''seeing'' more classes during training. In addition, when we increase our database size by 10 fold from 1000 to 10,000 (comparing table \ref{tab:embed} to \ref{tab:embed10k}), the accuracy does not drop by much showing that our method is relatively robust to the size of the database. 

Table \ref{tab:embedSize} repeats the embedding experiment comparing different embedding layer sizes on the 100 class LSTM model. Embedding quality increased from 128 to 256, but decreased from 256 to 512, suggesting that if the embedding becomes too large, overfitting on the training data reduces the ability to embed unknown classes.

\begin{figure}\centering
\frame{\includegraphics[width=\linewidth]{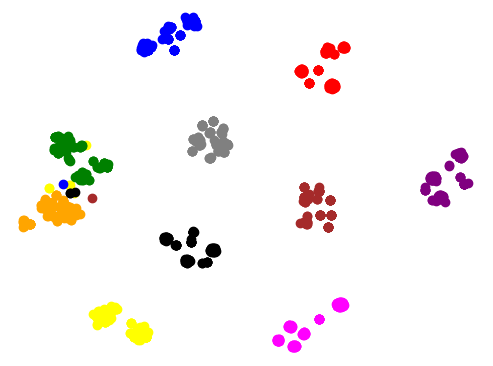}}
\caption{t-SNE visualization of embedded sequences from 10 protein classes using our 100-class model, each in a different color.}
\label{fig:tsne}
\end{figure}

In order to visualize our embedding space, we use t-SNE from Scikit-learn \cite{scikit-learn}, to transform 100 length-256 embeddings from each of 10 classes into a 2-dimensional space, where close vectors in the 256-dimensional space are also close in the 2-dimensional space. The embeddings were generated by the 100-class model using sequences from the test data, from 10 classes that were also used in training. The result in Figure \ref{fig:tsne} reveals that our embedding can in fact place similar sequences close together.

\begin{table}
\centering
\begin{tabular}{l|llll}
Model & N=1 & N=10 & N=20 & N=50 \\\hline
30 class LSTM & .252 & .423 & .488 & .591 \\
100 class LSTM & .350 & .598 & .680 & .793 \\
1000 class LSTM & .530 & .715 & .744 & .790 \\
100 class DSPACE & .231 & .360 & .416 & .502
\end{tabular}
\caption{\label{tab:embed}Fraction of query sequences with a correct match in the top N matches for each model, out of 1000 class pairs.}
\end{table}

\begin{table}
\centering
\begin{tabular}{l|llll}
Model & N=1 & N=10 & N=20 & N=50 \\\hline
30 class LSTM & .190 & .272 & .301 & .364 \\
100 class LSTM & .253 & .378 & .434 & .526 \\
1000 class LSTM & .389 & .592 & .641 & .701 \\
100 class DSPACE & .202 & .273 & .299 & .346
\end{tabular}
\caption{\label{tab:embed10k}Fraction of query sequences with a correct match in the top N matches for each model, out of 10000 class pairs.}
\end{table}

\begin{table}
\centering
\begin{tabular}{l|llll}
Embedding Size & N=1 & N=10 & N=20 & N=50 \\\hline
128 & .338 & .591 & .674 & .767 \\
256 & .350 & .598 & .680 & .793 \\
512 & .326 & .508 & .585 & .671
\end{tabular}
\caption{\label{tab:embedSize}Fraction of query sequences with a correct match in the top N matches for each model, out of 1000 class pairs. Comparison of different embedding sizes using the 100-class LSTM model.}
\end{table}

\subsection{Length Analysis}

\begin{figure*}\centering
\includegraphics[width=\linewidth]{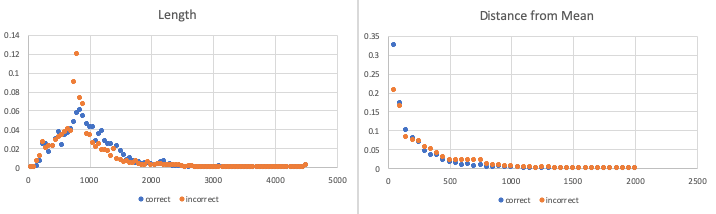}
\caption{The probability distributions of sequence length (left) and distance from mean sequence length (right) for correct and incorrect examples (each adds up to 1). The fact that the percentage of correct vs. incorrect classifications stays relatively constant, shows that our network is not relying heavily on the sequence length.}
\label{fig:lengths}
\end{figure*}

One of our concerns was that the classifier was making decisions simply based on the length of each sequence rather than its content. Therefore, we compared the lengths of sequences correctly classified by the 100-class LSTM model to the lengths of incorrectly classified sequences in the test set with 1 million examples. The results in Figure \ref{fig:lengths} indicate that while correct sequences are more likely to be close to the mean, for a given difference in length the accuracy does not change much. Therefore, length is not an important factor in the classifier's decision.

\subsection{Noise Analysis}
\label{subsec:noise}

\begin{figure}\centering
\includegraphics[width=\linewidth]{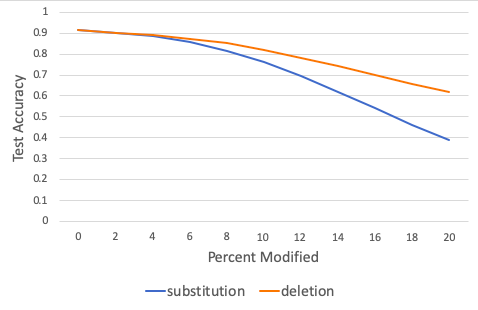}
\caption{The accuracy of our classification model under different noise sources as a function of the amount of noise. See Sec. \ref{subsec:noise} for more details.}
\label{fig:mutation}
\end{figure}

To test how robust our model is to noise and mutations in a DNA sequence, we performed alterations on each sequence in the 100-class test set and examined how they effect test accuracy. We tried two noise sources:
\begin{itemize}
    \item A simple probabilistic model of read noise from \cite{6620505}. The basic idea is that we go through the sequence and change each base independently with a given probability $p$.
    \item We also look at how the network behaves when part of the sequence is missing. We randomly select a starting base from our sequence and then remove $p$ percent of the total sequence length.
\end{itemize}

 Figure \ref{fig:mutation} shows how accuracy decreases as a function of $p$ for both noise sources. As expected, large mutations harm accuracy, but overall the model is able to handle small mutations without significant error.

One thing to note is that when removing parts of the sequence we chose a starting base at a location which is divisible by 3. Since our model depends on converting groups of 3 nucleotides to an amino acid, accuracy for deletions decreased significantly if the deleted segment was not aligned by groups of 3. For example, removing a single nucleotide from the sequence will cause all amino acid encodings after that nucleotide to be changed. 

A simple way around this issue in the future is to use amino acid sequences instead of nucleotide sequences. Another possibility is to change the stride on the first layer of our network from 3 to 1, so that every possible group of 3 nucleotides is considered. When we tried training with stride 1, test accuracy decreased slightly and training took significantly longer, so although this approach is costly it is possibly helpful if this type of noise is a concern.

\section{Conclusion}
\label{sec:conclusion}

In this work we presented a novel method for classifying and embedding DNA sequences. Using this method in conjunction with fast nearest neighbor algorithms we can find best matches to a query sequence even if it is from a previously unseen class. More importantly, using this embedding space provides not only a simple best match, but also distances to other sequences, thus providing functional information even for sequences which do not have an exact match in our database. We measure the robustness of our method both to dealing with unseen classes and to dealing with different noise sources.

We expect this type of work to be useful for many applications besides simple gene annotation. For example, we have begun working on using these embeddings to characterize microbial biogeographic provinces in the ocean. The basic idea is to embed ocean sequence data collected from different regions and examine if certain regions are more similar in our embedding space than others. This problem would be hard to solve using traditional comparative genomics methods as many of the genes are unknown and it would be very expensive to do a comprehensive BLAST comparison. Our initial results show that these embeddings lead to plausible groupings of regions.

Although we show promising results in this paper, more work needs to be done to understand the full potential from our method. For example, as described in Sec. \ref{subsec:class_results}, given the dataset used, accuracy is not expected to be perfect because of the ground truth classes given. The class labels used for training the classifier are not mutually exclusive and therefore confusion between such classes is expected due to the overlap. This is a difficult biological problem, because experimental determination of gene function is labor intensive \cite{michalska2015new} and virtually all gene databases are polluted by annotation errors \cite{green2005genome, jones2007estimating, schnoes2009annotation}. More accurate results will require non-overlapping, error-free databases. 

In addition, 1000 random classes (the largest number of classes we used for training) might not be representative enough of the entire database. More work can be done both in examining the effect of using larger amounts of classes for training and how to select a diverse set of classes to better represent the entire genome. 

Finally, we are currently further examining the parameter space and architecture changes which will lead to better embeddings. For example, instead of training using a classification layer we are experimenting with using triplet-loss \cite{schroff2015facenet} to train the embedding layer directly. This has led to improved embeddings for face recognition and therefore we expect to see an improvement in our embeddings as well.

However, even with these limitations, the results presented in this work reveal that our embedding is indeed capable of placing sequences with similar function close together, even when that class is not seen in training. Beyond simply speeding up the search for similar proteins, an embedding could allow prediction of protein function and properties, and lead to new ways of using sequence data in biology research. 

\section{Acknowledgements}
This material is based upon work supported by (1) the University of Tennessee, Knoxville College of Arts and Sciences, (2) Tickle College of Engineering, and (3) the Joint Institute for Computational Sciences. Any opinions, findings, conclusions, or recommendations expressed in this material are those of the author(s) and do not necessarily reflect the views of the University of Tennessee or Intel Corporation. 

\textcopyright 2019 IEEE.  Personal use of this material is permitted.  Permission from IEEE must be obtained for all other uses, in any current or future media, including reprinting/republishing this material for advertising or promotional purposes, creating new collective works, for resale or redistribution to servers or lists, or reuse of any copyrighted component of this work in other works.

\bibliographystyle{./bibliography/IEEEtran}
\bibliography{./bibliography/ref}

\begin{thebibliography}{10}
\providecommand{\url}[1]{#1}
\csname url@samestyle\endcsname
\providecommand{\newblock}{\relax}
\providecommand{\bibinfo}[2]{#2}
\providecommand{\BIBentrySTDinterwordspacing}{\spaceskip=0pt\relax}
\providecommand{\BIBentryALTinterwordstretchfactor}{4}
\providecommand{\BIBentryALTinterwordspacing}{\spaceskip=\fontdimen2\font plus
\BIBentryALTinterwordstretchfactor\fontdimen3\font minus
  \fontdimen4\font\relax}
\providecommand{\BIBforeignlanguage}[2]{{%
\expandafter\ifx\csname l@#1\endcsname\relax
\typeout{** WARNING: IEEEtran.bst: No hyphenation pattern has been}%
\typeout{** loaded for the language `#1'. Using the pattern for}%
\typeout{** the default language instead.}%
\else
\language=\csname l@#1\endcsname
\fi
#2}}
\providecommand{\BIBdecl}{\relax}
\BIBdecl

\bibitem{altschul1990basic}
S.~F. Altschul, W.~Gish, W.~Miller, E.~W. Myers, and D.~J. Lipman, ``Basic
  local alignment search tool,'' \emph{Journal of molecular biology}, vol. 215,
  no.~3, pp. 403--410, 1990.

\bibitem{rost2002enzyme}
B.~Rost, ``Enzyme function less conserved than anticipated,'' \emph{Journal of
  molecular biology}, vol. 318, no.~2, pp. 595--608, 2002.

\bibitem{pesant2015open}
S.~Pesant, F.~Not, M.~Picheral, S.~Kandels-Lewis, N.~Le~Bescot, G.~Gorsky,
  D.~Iudicone, E.~Karsenti, S.~Speich, R.~Troubl{\'e} \emph{et~al.}, ``Open
  science resources for the discovery and analysis of tara oceans data,''
  \emph{Scientific data}, vol.~2, p. 150023, 2015.

\bibitem{Sunagawa2015structure}
S.~Sunagawa, L.~P. Coelho, S.~Chaffron, J.~R. Kultima, K.~Labadie, G.~Salazar,
  B.~Djahanschiri, G.~Zeller, D.~R. Mende, A.~Alberti \emph{et~al.},
  ``Structure and function of the global ocean microbiome,'' \emph{Science},
  vol. 348, no. 6237, p. 1261359, 2015.

\bibitem{steen2019high}
A.~D. Steen, A.~Crits-Christoph, P.~Carini, K.~M. DeAngelis, N.~Fierer, K.~G.
  Lloyd, and J.~C. Thrash, ``High proportions of bacteria and archaea across
  most biomes remain uncultured,'' \emph{The ISME journal}, pp. 1--5, 2019.

\bibitem{lloyd2018phylogenetically}
K.~G. Lloyd, A.~D. Steen, J.~Ladau, J.~Yin, and L.~Crosby, ``Phylogenetically
  novel uncultured microbial cells dominate earth microbiomes,''
  \emph{MSystems}, vol.~3, no.~5, pp. e00\,055--18, 2018.

\bibitem{oleary2016refseq}
N.~A. O'Leary, M.~W. Wright, J.~R. Brister, S.~Ciufo, D.~Haddad, R.~McVeigh,
  B.~Rajput, B.~Robbertse, B.~Smith-White, D.~Ako-Adjei, A.~Astashyn,
  A.~Badretdin, Y.~Bao, O.~Blinkova, V.~Brover, V.~Chetvernin, J.~Choi, E.~Cox,
  O.~D. Ermolaeva, C.~M. Farrell, T.~Goldfarb, T.~Gupta, D.~H. Haft,
  E.~Hatcher, W.~Hlavina, V.~S. Joardar, V.~K. Kodali, W.~Li, D.~R. Maglott,
  P.~Masterson, K.~M. McGarvey, M.~R. Murphy, K.~O'Neill, S.~Pujar, S.~H.
  Rangwala, D.~Rausch, L.~D. Riddick, C.~L. Schoch, A.~Shkeda, S.~S. Storz,
  H.~Sun, F.~Thibaud-Nissen, I.~Tolstoy, R.~E. Tully, A.~R. Vatsan, C.~Wallin,
  D.~Webb, W.~Wu, M.~J. Landrum, A.~Kimchi, T.~A. Tatusova, M.~DiCuccio, P.~A.
  Kitts, T.~D. Murphy, and K.~D. Pruitt, ``Reference sequence (refseq) database
  at ncbi: current status, taxonomic expansion, and functional annotation.''
  \emph{Nucleic Acids Research}, vol.~44, no. Database-Issue, pp. 733--745,
  2016.

\bibitem{altschul1997gapped}
\BIBentryALTinterwordspacing
S.~F. Altschul, T.~L. Madden, A.~A. Schäffer, J.~Zhang, Z.~Zhang, W.~Miller,
  and D.~J. Lipman, ``{Gapped BLAST and PSI-BLAST: a new generation of protein
  database search programs},'' \emph{Nucleic Acids Research}, vol.~25, no.~17,
  pp. 3389--3402, 09 1997. [Online]. Available:
  \url{https://doi.org/10.1093/nar/25.17.3389}
\BIBentrySTDinterwordspacing

\bibitem{camacho2009blast+}
C.~Camacho, G.~Coulouris, V.~Avagyan, N.~Ma, J.~Papadopoulos, K.~Bealer, and
  T.~L. Madden, ``Blast+: architecture and applications,'' \emph{BMC
  bioinformatics}, vol.~10, no.~1, p. 421, 2009.

\bibitem{edgar2010usearch}
\BIBentryALTinterwordspacing
R.~C. Edgar, ``{Search and clustering orders of magnitude faster than BLAST},''
  \emph{Bioinformatics}, vol.~26, no.~19, pp. 2460--2461, 08 2010. [Online].
  Available: \url{https://doi.org/10.1093/bioinformatics/btq461}
\BIBentrySTDinterwordspacing

\bibitem{buchfink2015fast}
B.~Buchfink, C.~Xie, and D.~H. Huson, ``Fast and sensitive protein alignment
  using diamond,'' \emph{Nature methods}, vol.~12, no.~1, p.~59, 2015.

\bibitem{vouzis2010gpu}
P.~D. Vouzis and N.~V. Sahinidis, ``Gpu-blast: using graphics processors to
  accelerate protein sequence alignment,'' \emph{Bioinformatics}, vol.~27,
  no.~2, pp. 182--188, 2010.

\bibitem{sawyer2015hpc}
S.~E. Sawyer, B.~Rekepalli, M.~D. Horton, and R.~G. Brook, ``Hpc-blast:
  distributed blast for xeon phi clusters,'' in \emph{Proceedings of the 6th
  ACM Conference on Bioinformatics, Computational Biology and Health
  Informatics}.\hskip 1em plus 0.5em minus 0.4em\relax ACM, 2015, pp. 512--513.

\bibitem{ye2017hblast}
\BIBentryALTinterwordspacing
W.~Ye, Y.~Chen, Y.~Zhang, and Y.~Xu, ``{H-BLAST: a fast protein sequence
  alignment toolkit on heterogeneous computers with GPUs},''
  \emph{Bioinformatics}, vol.~33, no.~8, pp. 1130--1138, 01 2017. [Online].
  Available: \url{https://doi.org/10.1093/bioinformatics/btw769}
\BIBentrySTDinterwordspacing

\bibitem{muja2009fast}
M.~Muja and D.~G. Lowe, ``Fast approximate nearest neighbors with automatic
  algorithm configuration.'' \emph{VISAPP (1)}, vol.~2, no. 331-340, p.~2,
  2009.

\bibitem{iwasaki2018optimization}
M.~Iwasaki and D.~Miyazaki, ``Optimization of indexing based on k-nearest
  neighbor graph for proximity search in high-dimensional data,'' \emph{arXiv
  preprint arXiv:1810.07355}, 2018.

\bibitem{eddy2011accelerated}
S.~R. Eddy, ``Accelerated profile hmm searches,'' \emph{PLoS computational
  biology}, vol.~7, no.~10, p. e1002195, 2011.

\bibitem{seemann2014prokka}
T.~Seemann, ``Prokka: rapid prokaryotic genome annotation,''
  \emph{Bioinformatics}, vol.~30, no.~14, pp. 2068--2069, 2014.

\bibitem{haft2012tigrfams}
D.~H. Haft, J.~D. Selengut, R.~A. Richter, D.~Harkins, M.~K. Basu, and E.~Beck,
  ``Tigrfams and genome properties in 2013,'' \emph{Nucleic acids research},
  vol.~41, no.~D1, pp. D387--D395, 2012.

\bibitem{woloszynek201916s}
S.~Woloszynek, Z.~Zhao, J.~Chen, and G.~L. Rosen, ``16s rrna sequence
  embeddings: Meaningful numeric feature representations of nucleotide
  sequences that are convenient for downstream analyses,'' \emph{PLoS
  computational biology}, vol.~15, no.~2, p. e1006721, 2019.

\bibitem{schwartz2018deep}
A.~S. Schwartz, G.~J. Hannum, Z.~R. Dwiel, M.~E. Smoot, A.~R. Grant, J.~M.
  Knight, S.~A. Becker, J.~R. Eads, M.~C. LaFave, H.~Eavani \emph{et~al.},
  ``Deep semantic protein representation for annotation, discovery, and
  engineering,'' \emph{BioRxiv}, p. 365965, 2018.

\bibitem{quang2016danq}
D.~Quang and X.~Xie, ``Danq: a hybrid convolutional and recurrent deep neural
  network for quantifying the function of dna sequences,'' \emph{Nucleic acids
  research}, vol.~44, no.~11, pp. e107--e107, 2016.

\bibitem{1baum1967inequality}
L.~E. Baum and J.~A. Eagon, ``An inequality with applications to statistical
  estimation for probabilistic functions of markov processes and to a model for
  ecology,'' \emph{Bulletin of the American Mathematical Society}, vol.~73,
  no.~3, pp. 360--363, 1967.

\bibitem{2viterbi1967error}
A.~Viterbi, ``Error bounds for convolutional codes and an asymptotically
  optimum decoding algorithm,'' \emph{IEEE transactions on Information Theory},
  vol.~13, no.~2, pp. 260--269, 1967.

\bibitem{3rabiner1989tutorial}
L.~R. Rabiner, ``A tutorial on hidden markov models and selected applications
  in speech recognition,'' \emph{Proceedings of the IEEE}, vol.~77, no.~2, pp.
  257--286, 1989.

\bibitem{4lukashin1998genemark}
A.~V. Lukashin and M.~Borodovsky, ``Genemark. hmm: new solutions for gene
  finding,'' \emph{Nucleic acids research}, vol.~26, no.~4, pp. 1107--1115,
  1998.

\bibitem{5sutskever2011generating}
I.~Sutskever, J.~Martens, and G.~E. Hinton, ``Generating text with recurrent
  neural networks,'' in \emph{Proceedings of the 28th International Conference
  on Machine Learning (ICML-11)}, 2011, pp. 1017--1024.

\bibitem{6hochreiter1997long}
S.~Hochreiter and J.~Schmidhuber, ``Long short-term memory,'' \emph{Neural
  computation}, vol.~9, no.~8, pp. 1735--1780, 1997.

\bibitem{7graves2013generating}
A.~Graves, ``Generating sequences with recurrent neural networks,'' \emph{arXiv
  preprint arXiv:1308.0850}, 2013.

\bibitem{schuster1997bidirectional}
M.~Schuster and K.~K. Paliwal, ``Bidirectional recurrent neural networks,''
  \emph{IEEE Transactions on Signal Processing}, vol.~45, no.~11, pp.
  2673--2681, 1997.

\bibitem{mikolov2013efficient}
T.~Mikolov, K.~Chen, G.~Corrado, and J.~Dean, ``Efficient estimation of word
  representations in vector space,'' \emph{arXiv preprint arXiv:1301.3781},
  2013.

\bibitem{parkhi2015deep}
O.~M. Parkhi, A.~Vedaldi, A.~Zisserman \emph{et~al.}, ``Deep face
  recognition.'' in \emph{bmvc}, vol.~1, no.~3, 2015, p.~6.

\bibitem{ZHANG20151857}
\BIBentryALTinterwordspacing
D.~Zhang, H.~Xu, Z.~Su, and Y.~Xu, ``Chinese comments sentiment classification
  based on word2vec and svmperf,'' \emph{Expert Systems with Applications},
  vol.~42, no.~4, pp. 1857 -- 1863, 2015. [Online]. Available:
  \url{http://www.sciencedirect.com/science/article/pii/S0957417414005508}
\BIBentrySTDinterwordspacing

\bibitem{sadovnik2018finding}
A.~Sadovnik, W.~Gharbi, T.~Vu, and A.~Gallagher, ``Finding your lookalike:
  Measuring face similarity rather than face identity,'' in \emph{Proceedings
  of the IEEE Conference on Computer Vision and Pattern Recognition Workshops},
  2018, pp. 2345--2353.

\bibitem{pearson2003finding}
W.~Pearson, ``Finding protein and nucleotide similarities with fasta,''
  \emph{Current protocols in bioinformatics}, vol.~4, no.~1, pp. 3--9, 2003.

\bibitem{scikit-learn}
F.~Pedregosa, G.~Varoquaux, A.~Gramfort, V.~Michel, B.~Thirion, O.~Grisel,
  M.~Blondel, P.~Prettenhofer, R.~Weiss, V.~Dubourg, J.~Vanderplas, A.~Passos,
  D.~Cournapeau, M.~Brucher, M.~Perrot, and E.~Duchesnay, ``Scikit-learn:
  Machine learning in {P}ython,'' \emph{Journal of Machine Learning Research},
  vol.~12, pp. 2825--2830, 2011.

\bibitem{6620505}
A.~{Motahari}, K.~{Ramchandran}, D.~{Tse}, and N.~{Ma}, ``Optimal dna shotgun
  sequencing: Noisy reads are as good as noiseless reads,'' in \emph{2013 IEEE
  International Symposium on Information Theory}, July 2013, pp. 1640--1644.

\bibitem{michalska2015new}
K.~Michalska, A.~D. Steen, G.~Chhor, M.~Endres, A.~T. Webber, J.~Bird, K.~G.
  Lloyd, and A.~Joachimiak, ``New aminopeptidase from “microbial dark
  matter” archaeon,'' \emph{The FASEB Journal}, vol.~29, no.~9, pp.
  4071--4079, 2015.

\bibitem{green2005genome}
M.~Green and P.~Karp, ``Genome annotation errors in pathway databases due to
  semantic ambiguity in partial ec numbers,'' \emph{Nucleic acids research},
  vol.~33, no.~13, pp. 4035--4039, 2005.

\bibitem{jones2007estimating}
C.~E. Jones, A.~L. Brown, and U.~Baumann, ``Estimating the annotation error
  rate of curated go database sequence annotations,'' \emph{BMC
  bioinformatics}, vol.~8, no.~1, p. 170, 2007.

\bibitem{schnoes2009annotation}
A.~M. Schnoes, S.~D. Brown, I.~Dodevski, and P.~C. Babbitt, ``Annotation error
  in public databases: misannotation of molecular function in enzyme
  superfamilies,'' \emph{PLoS computational biology}, vol.~5, no.~12, p.
  e1000605, 2009.

\bibitem{schroff2015facenet}
F.~Schroff, D.~Kalenichenko, and J.~Philbin, ``Facenet: A unified embedding for
  face recognition and clustering,'' in \emph{Proceedings of the IEEE
  conference on computer vision and pattern recognition}, 2015, pp. 815--823.

\end{thebibliography}
\end{document}